\documentclass[runningheads]{llncs}
\usepackage[T1]{fontenc}
\usepackage{graphicx}
\usepackage{booktabs}
\usepackage[misc]{ifsym}
\usepackage{cite}
\usepackage{amsmath,amssymb,amsfonts}
\usepackage{array}
\usepackage{tabularx}
\usepackage{algorithmic}
\usepackage{graphicx}
\usepackage{subcaption}
\usepackage{textcomp}
\usepackage{xcolor}
\usepackage{printlen}
\usepackage{multirow}
\newcommand{\euler}{e}

\DeclareMathOperator*{\argmin}{arg\,min}

\def\BibTeX{{\rm B\kern-.05em{\sc i\kern-.025em b}\kern-.08em
    T\kern-.1667em\lower.7ex\hbox{E}\kern-.125emX}}
\newcommand{\corr}{(\Letter)}
\usepackage{mwe}

\begin{document}

\title{The Significance of Latent Data Divergence in
Predicting System Degradation}

\titlerunning{Latent Data Divergence in
Predicting System Degradation}

\author{Miguel Fernandes\inst{1}\orcidID{0000-0002-7800-6854} \corr \and
Catarina Silva\inst{1}\orcidID{0000-0002-5656-0061} \and
Alberto Cardoso \inst{1}\orcidID{0000-0003-1824-1075} \and
Bernardete Ribeiro \inst{1}\orcidID{0000-0002-9770-7672}}


\institute{Centre for Informatics and Systems, Department of Informatics Engineering of the University of Coimbra, University of Coimbra, Coimbra, Portugal \email{\{mfernandes,catarina,alberto,bribeiro\}dei.uc.pt}}

\maketitle              

\begin{abstract}
Condition-Based Maintenance is pivotal in enabling the early detection of potential failures in engineering systems, where precise prediction of the Remaining Useful Life is essential for effective maintenance and operation. However, a predominant focus in the field centers on predicting the Remaining Useful Life using unprocessed or minimally processed data, frequently neglecting the intricate dynamics inherent in the dataset. In this work we introduce a novel methodology grounded in the analysis of statistical similarity within latent data from system components. Leveraging a specifically designed architecture based on a Vector Quantized Variational Autoencoder, we create a sequence of discrete vectors which is used to estimate system-specific priors. We infer the similarity between systems by evaluating the divergence of these priors, offering a nuanced understanding of individual system behaviors. The efficacy of our approach is demonstrated through  experiments on the NASA commercial modular aero-propulsion system simulation (C-MAPSS) dataset. Our validation not only underscores the potential of our method in advancing the study of latent statistical divergence but also demonstrates its superiority over existing techniques.

\end{abstract}

\section{Introduction}

Condition-Based Maintenance (CBM) has become crucial in various industries, involving real-time monitoring of system conditions through a range of sensors and data acquisition techniques \cite{cbm_introduction}. By continuously collecting and analyzing operational data, CBM enables the early detection of potential failures, providing valuable insights into the Remaining Useful Life (RUL) of equipment \cite{cbm_jardine}. A significant advantage of CBM is its capacity to transition maintenance practices from a reactive to a proactive and predictive approach, preventing downtime and costly repairs associated with traditional reactive maintenance.

Most RUL prediction techniques in the literature often utilize deep neural networks such as CNNs \cite{CNN_Babu, DCNN_Li, MS_DCNN_Li}, LSTMs \cite{LSTM_FNN_Zheng, LSTM_xia}, GRUs \cite{Attention_Liu, GRU_Attention_Liu}, hybrid models \cite{HDNN_Al, Attention_Liu, MTSTAN_Li, LSTM_MLSA_Xia}, and attention mechanisms \cite{Attention_Liu, GRU_Attention_Liu, LSTM_MLSA_Xia}, to process raw or minimally processed data for forecasting system degradation \cite{CNN_Babu, DCNN_Li, MS_DCNN_Li, LSTM_FNN_Zheng, LSTM_xia, Attention_Liu, GRU_Attention_Liu, HDNN_Al, MTSTAN_Li, LSTM_MLSA_Xia}. While these approaches are useful, they may encounter challenges in handling non-linear relationships between data and RUL, especially in systems with intricate degradation patterns, which can lead to performance issues and limited generalizability across various operational conditions.

Another avenue in RUL estimation research involves the use of Autoencoders that transform input data into a latent representation, thereby facilitating the extraction of features and patterns crucial for accurate RUL prediction \cite{Autoencoder_Andre, Autoencoder_Gan_Hou, vae_Qin_hi, rnn_auto_hi_yu, RVE_Costa}. Ellefsen et al. \cite{Autoencoder_Andre} and Hou et al.  \cite{Autoencoder_Gan_Hou} have proposed methods where the data is initially processed into a latent representation and then used by an LSTM network to predict the RUL. Instead, Qin et al. \cite{vae_Qin_hi} and Yu et al. \cite{rnn_auto_hi_yu} create Health Indicators using latent representations from autoencoders which are used as input for RUL prediction using either an LSTM \cite{vae_Qin_hi} or a similarity-based approach \cite{rnn_auto_hi_yu}. 

Despite recent advancements, a considerable gap persists in comprehending the statistical variations within the latent space and their correlation with real-world degradation patterns. Our research addresses this gap by thoroughly exploring the latent space, emphasizing its statistical dynamics and their interplay with degradation processes, ultimately providing a deeper insight into the latent data structures for an enhanced understanding of system health and degradation. 

Our key contributions include: (1) a novel architecture using Transformer \cite{Transformer} and Vector Quantized Variational Autoencoder's (VQ-VAE) \cite{VQVAE} components, generating a sequence of discrete vectors as the latent block of the model - this distinctive sequence of discrete vectors represents an unexplored approach in existing literature; (2) a novel methodology for modeling the latent space of the data using transition matrices, enabling the computation of steady-state distributions for the efficient estimation of system-specific priors; (3) a new methodology to evaluate statistical divergence in priors, offering a robust framework for assessing data similarity and predicting the RUL of a system; and (4) the validation and performance comparison using the C-MAPSS dataset \cite{cmapps}, demonstrating superior performance compared to other state-of-the-art methods. This validation underscores the significance of gaining a deeper insight into latent data structures for an improved understanding of a system's health and degradation.

The remainder of the document is structured as follows: section 2 presents the problem statement, section 3 presents the preliminaries necessary for understanding our work, section 4 presents the proposed approach, section 5 presents the experiments and section 6 presents the conclusion and future work.

\section{Problem Statement}

We first define notations used throughout this paper. The dataset is represented as $D = \{x_{1}, x_{2}, ..., x_{N}\}$, consisting of $N$ instances. Each sequence $x_i$ in D is a time series represented as $x_i = \{x_i^1, x_i^2, ..., x_i^{T_i}\}$, where $x_i^t$ denotes the observation at time $t$ within sequence $i$, and $T_i$ is the length of sequence $i$. Furthermore, we introduce multiple subsets within $D$ denoted as $D_s$, each representing a specific system or a segment of the data relevant to a particular context. This subset $D_s$ may include one or several sequences from $D$. The distinction between $D$ and its subsets $D_s$ is crucial for our analysis, allowing us to focus on individual systems or data segments, while enabling inferences and comparisons across the broader dataset.

In the context of machine learning, the objective is to model and approximate the true, unknown data distribution $p_{*}(x)$, which is inherently time-dependent. To achieve this, we approximate this underlying process with chosen model $p_{\theta}(x)$ with parameters $\theta$, such that $p_{\theta}(x) \approx p_{*}(x)$

\subsection{Latent Variables}

Latent variables, represented as 
$z$, are critical components of our model's architecture. These variables capture underlying features and patterns within the data that are not directly observable or explicitly represented in the dataset. Their role is to internally represent the complex characteristics of the data \cite{VAEintro, VAE}. In the context of this scenario, $p_{\theta}(x)$ can be expressed as the sum over all possible latent variables $z$:
$$p_{\theta}(x) = \sum_{z \in Z} p_{\theta}(x|z) p_{\theta}(z)$$ 
It is important to note that, for the purposes of this specific work, we operate under the assumption that the $x$ space and the latent space $Z$ are discrete.
Furthermore, $p_{\theta}(x|z)$ is denoted as the decoder whose role is to reconstruct the input data from $z$. It essentially translates the encoded, latent data back into a form that closely approximates the original input data, thereby learning the conditional distribution of the data given the latent variables. On the other hand, the encoder, termed $q_{\sigma}(z|x)$, serves as an approximation of the intractable true posterior $p_{\theta}(z|x)$. The encoder's function is to map the input data into the latent space, providing a probabilistic representation of the input in terms of the latent variables.

\subsection{Statistical Divergence}

One prominent method employed in measuring the divergence between probability distributions is the \textit{Jensen–Shannon} divergence, $JS$ \cite{information_Cover}. Unlike other measures, such as the \textit{Kullback-Leibler} divergence $KL$, $JS$ divergence provides a symmetric evaluation of the difference between two distributions. This symmetry is particularly advantageous for our purposes because it ensures that the divergence is independent of the order of the distributions. This property makes $JS$ divergence well-suited for applications where the direction of divergence is not a primary concern, facilitating a more balanced and equitable comparison between the distributions. The $JS$ divergence is given by:
\begin{equation*}
    JS(p(x) \| q(x)) = \frac{1}{2}KL \left( p(x) \Big\| \frac{p(x)+q(x)}{2} \right)
    + \frac{1}{2}KL \left( q(x) \Big\| \frac{p(x)+q(x)}{2} \right),
\end{equation*}
where the $KL$ divergence is given by:
\begin{equation*}
    KL(p(x) \| q(x)) = \sum_{x \in D} p(x) \log \left( \frac{p(x)}{q(x)} \right)  
\end{equation*}
In our work, we leverage the $JS$ divergence between data as a valuable tool for quantifying and understanding the similarity between two probability distributions.

\subsection{Objective}

The main objective of our research is to identify, the subset $D_{s}$ within a dataset $D$ that most closely resembles a specific subset under study $D_j$. Our approach is grounded in the hypothesis that systems with similar data subsets, indicating comparable operational behaviors and degradation trajectories, are likely to exhibit similar RULs. This objective can be formulated as follows:
$$\argmin_{D_s \in D} JS(p_{*}(x|D=D_{j})||p_{*}(x|D=D_{s}))$$
We aim to find the subset $D_{s}$ within $D$, where $j \neq s$, that minimizes the $JS$ divergence from our subset under study $D_j$. Making the assumption that $p_{\theta}(x)$ serves as a perfect approximator of $p_{*}(x)$, $JS(p_{\theta}(x|D=D_j)||p_{\theta}(x|D=D_s))$ is given by: 
\begin{multline*}
\frac{1}{2} \left[ \sum_{x \in D} p_{\theta}(x|D=D_{j}) \log \left( \frac{p_{\theta}(x|D=D_{j})}{\frac{p_{\theta}(x|D=D_{j}) + p_{\theta}(x|D=D_{s})}{2}} \right) \right.  \\
+ \left. \sum_{x \in D} p_{\theta}(x|D=D_{s}) \log \left( \frac{p_{\theta}(x|D=D_{s})}{\frac{p_{\theta}(x|D=D_{j}) + p_{\theta}(x|D=D_{s})}{2}} \right) \right]
\end{multline*}
During training, assuming the decoder is conditioned on $z = z_{q}(x)$, the latent representation derived from the input data $x$, implies that the decoder should not allocate any probability mass to $p_{\theta}(x|z)$ for $z \neq z_{q}(x)$ once it has fully converged \cite{VQVAE}. Thus we can write that $p_{\theta}(x) \approx p_{\theta}(x|z_{q}(x)) p_{\theta}(z_{q}(x))$. In addition, as the model is not conditioned on $D$ during training, we can reasonably assume that $z$ encapsulates all relevant information about $x$. Consequently, this allows us to approximate $p_{\theta}(x|D=D_{s}) \approx p_{\theta}(x|z_{q}(x)) p_{\theta}(z_{q}(x)|D=D_{s})$, leading to the following formulation:
\begin{multline*}
KL (p_{\theta}(x|D=D_{j}) \| p_{\theta}(x|D=D_{s})) =\\
\\ \sum_{x \in D} p_{\theta}(x|D=D_{j}) \log \left( \frac{p_{\theta}(z_{q}(x)|D=D_{j})}{\frac{p_{\theta}(z_{q}(x)|D=D_{j}) + p_{\theta}(z_{q}(x)|D=D_{s})}{2}} \right)
\end{multline*}
Since $p_{\theta}(x)$ is a perfect estimator of $p_{*}(x)$, and given that every data point $x$ in $D$ is mapped to its corresponding latent variable $z_{q}(x)$ in the latent space $Z$, then we can say that $\sum_{x \in D} p_{\theta}(x|z_{q}(x))p_{\theta}(z_{q}(x)|D=D_s) = \sum_{z \in Z} p_{\theta}(z|D=D_s)$. Therefore, we can reformulate the $JS$ divergence as:
\begin{multline*}
\frac{1}{2} \left[ \sum_{z \in Z} p_{\theta}(z|D=D_{j}) \log \left( \frac{p_{\theta}(z|D=D_{j})}{\frac{p_{\theta}(z|D=D_{j}) + p_{\theta}(z|D=D_{s})}{2}} \right) \right. \\
+ \left. \sum_{z \in Z} p_{\theta}(z|D=D_{s}) \log \left( \frac{p_{\theta}(z|D=D_{s})}{\frac{p_{\theta}(z|D=D_{j}) + p_{\theta}(z|D=D_{s})}{2}} \right) \right]
\end{multline*}
This implies that:
\begin{equation*}
JS(p(x|D=D_{j})\|p(x|D=D_{s})) \approx JS(p(z|D=D_{j})\|p(z|D=D_{s}))
\end{equation*}
Consequently, by minimizing the $JS$ divergence over the latent variables, we effectively approximate the minimization of the $JS$ divergence for $x$. While calculating the $JS$ divergence of latent variables might be more challenging than for $x$, subsequent sections of this paper will present a feasible approach to accomplish this task.

\section{Preliminaries}

In order to develop a prior that can be easily estimated and effectively utilized in our approach, a specific model structure is essential. To achieve this, we incorporate components from two main architectures: Transformers \cite{Transformer} and VQ-VAEs \cite{VQVAE}. These are critical in creating an architecture that facilitates the accurate transformation of raw data into a quantized latent space, suitable for the estimation of system priors. 

\subsection{Transformers}

Transformers \cite{Transformer} are models designed to handle sequential data such as time series. Their architecture deviates from traditional recurrent layers, instead relying exclusively on attention mechanisms to process data. Transformers follow an encoder-decoder architecture, where the encoder is responsible for processing and transforming the input data into a latent representation, whereas the decoder is tasked with taking the encoder’s representations and generating the final output. We briefly introduce the two pivotal components of the Transformer as following:

\subsubsection{Positional Encodings} Given that the architecture is devoid of any recurrent layers, introducing a method for acknowledging the sequence order is essential. To address this, positional encodings, $p$, are integrated into the input at the onset of both the encoder and decoder modules. These encodings have a similar dimension to the input, thereby these can be summed. The positional encodings are defined as follows:

\begin{equation*}
\begin{aligned}
p_{(pos, 2j)} = \sin\left(\frac{pos}{10000^{\frac{2j}{d}}}\right)
\\
p_{(pos, 2j+1)} = \cos\left(\frac{pos}{10000^{\frac{2j}{d}}}\right)
\end{aligned},
\end{equation*}
where $p_{(pos, j)}$ denotes the positional encoding for the element located at position $pos$ and dimension $j$. Here, $pos$ signifies the input's sequential position, ranging from $1$ to $N$, where $N$ is the total number of elements in the sequence. $j$ indicates the dimension index, which varies from $0$ to $\frac{d}{2} - 1$. The variable $d$ represents the dimensionality of the input. 

\subsubsection{Multi-Head Attention} In the Transformer architecture, capturing different facets of information within a sequence is paramount. To that end, the model employs a mechanism known as Multi-Head Attention. In this approach, multiple parallel attention heads are used to produce different output vectors. These outputs are then concatenated and linearly transformed to yield the final attention result. The Multi-Head Attention can be formalized as:
\begin{equation*}
\text{MultiHead}(Q, K, V) = \text{Concat}(\text{head}_1, \ldots, \text{head}_h) W^O,
\end{equation*}
where each individual head is computed as:
\begin{equation*}
\text{head}_i = \text{Attention}(Q W^Q_i, K W^K_i, V W^V_i)
\end{equation*}
\begin{equation*}
\text{Attention}
(Q, K, V) = \text{softmax} \left( \frac{QK^T}{\sqrt{d_k}} \right) V,
\end{equation*}
where $Q$, $K$, and $V$ are the input matrices for the query, key, and value, respectively. Specifically, $Q$ and $K$ have a dimensionality of $d_k$. $W^Q_i$, $W^K_i$, and $W^V_i$ are weight matrices corresponding to the $i$-th head for query, key, and value, respectively. $W^O$ is the final linear transformation weight matrix.

\subsection{VQ-VAE}

VQ-VAEs \cite{VQVAE, vqvae2} incorporate the concept of vector quantization, to enhance the modeling capabilities of VAEs \cite{VAE}. The key idea behind VQ-VAEs is to discretize the continuous latent space of VAEs into a set of discrete codes, which enables the model to avoid issues of "posterior collapse", where the latent representations are ignored.

In a VQ-VAE the encoder takes the input data and converts it into a continuous latent representation denoted as $z_e(x)$. This continuous representation is then quantized by mapping it to the nearest entry in a codebook $e \in \mathbb{R}^{K*N_{e}}$, where $K$ represents the size of the discrete latent space and $N_{e}$ is the number of discrete embeddings. This quantization process results in obtaining discrete states $z_q(x)$ that serve as the latent representation. These discrete states are then fed into the decoder to reconstruct the input data.

In the training process of VQ-VAEs, the objective function typically consists of three components: a reconstruction loss term that measures the difference between the input data and its reconstruction, a Vector Quantization term, which adjusts the embedding vectors $e$ towards the encoder outputs $z_e(x)$, and a commitment term to control the growth of the encoder outputs. Overall the training objective can be written as:
\begin{equation*}
    L = ||\hat{x} - x||_2 + ||sg(z_e(x)) - e||_2 + \beta||z_e(x) - sg(e)||_2,
\end{equation*}
where $sg$ denotes the stop-gradient operation, implying that the gradient is not considered during the optimization process, $\beta$ is a damping parameter, ensuring that the encoder's output commits to the embeddings without unbounded growth, and $\hat{x}$ represents the model's reconstruction of the input data.

\section{Proposed Approach}

In this section, we detail our approach to estimate the statistical divergence of priors for predicting the RUL of a system. Our method utilizes a novel model integrating Transformers \cite{Transformer} and Vector Quantizatized Autoecoder (VQVAE) \cite{VQVAE} components. This model efficiently generates a latent space that allows for straightforward estimation of priors using transition matrices. We will explain in detail how to derive these transition matrices from the latent space and demonstrate their utility in measuring the statistical divergence between different systems. By analyzing this divergence, we can ascertain which systems are most similar and use this insight to infer their RUL.

\subsection{Model Architecture}

In our approach, we focus on handling time series data. The input to our model is defined by $x_{s}^{t_s} \in \mathbb{R}^{T \times F}$, where $t_s$ serves as the identifier for a specific time window of system $s$. In the context of our problem statement, $x_{s}^{t_s}$ can be thought of as a sequence within the subset $D_s$. In this notation, $T$ represents the length of the time window, encompassing a certain number of sequential data points, while $F$ indicates the number of features. It is important to highlight that our approach utilizes overlapping time windows, meaning that consecutive time windows share common data points. This framework allows us to systematically analyze the time series data by capturing the temporal dynamics and feature interactions over the specified time window. Figure \ref{fig:Transformer_VQVAE} presents the proposed architecture that is comprised of three main components:
\begin{figure}[ht]
    \centering
    \includegraphics[width=175pt]{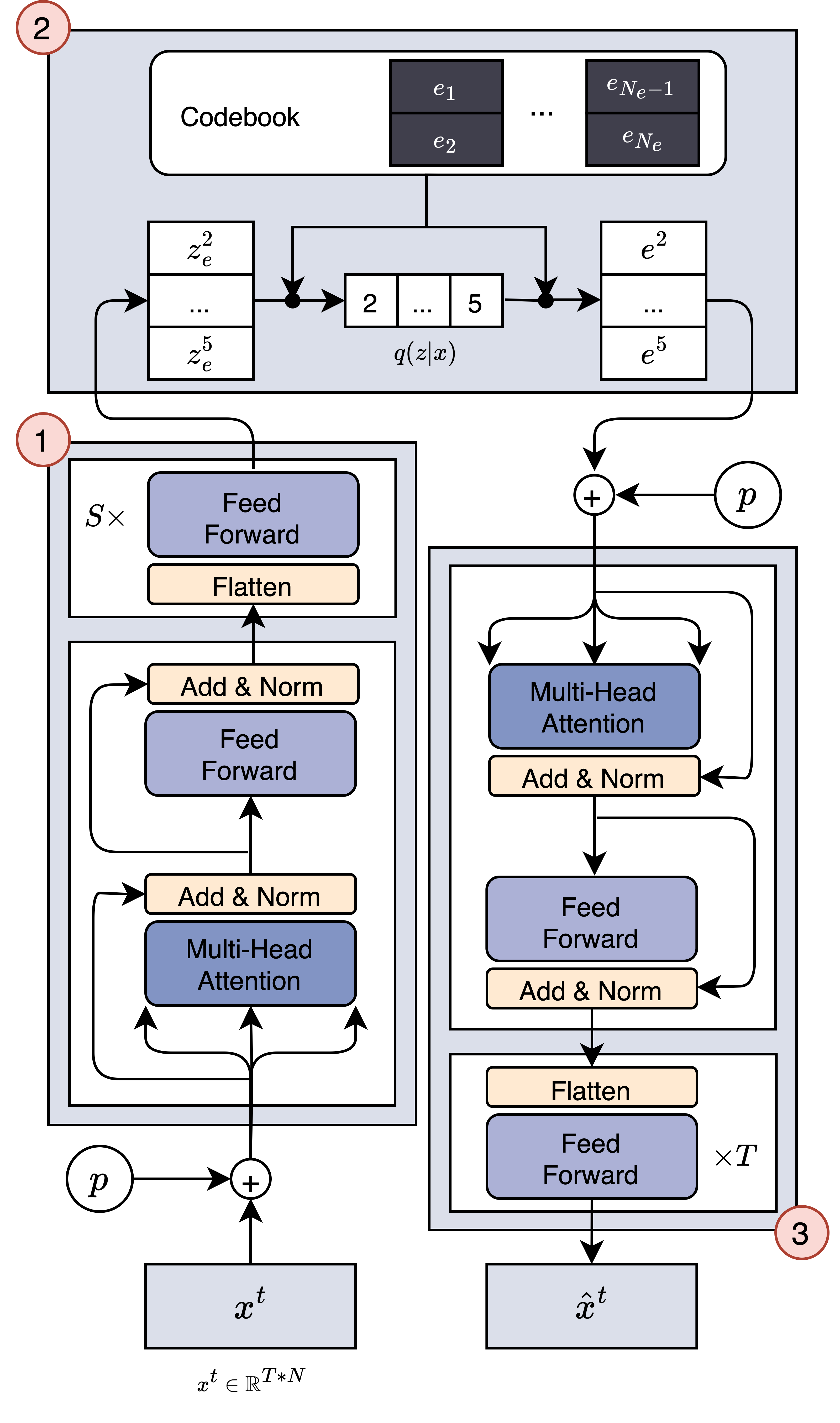}
    \caption{The architecture of the proposed model, consisting of three key components: 1) the encoder, responsible for processing the input data; 2) the vector quantization model, which discretizes the data into a latent representation; and 3) the decoder, tasked with reconstructing the output from the quantized data.}
    \label{fig:Transformer_VQVAE}
\end{figure}
\subsubsection{Encoder}

This component is tasked with generating a latent representation $z \in \mathbb{R}^{S \times E}$, where $S$ represents the number of sequences in the latent space and $E$ is the dimensionality of each sequence. In our design, the encoder is based on the Transformer architecture. The output of the Transformer's encoder is transformed to fit into the $S \times E$ dimensional space, using a series of Feed-forward Neural Networks (FNNs). These are integral in shaping the final latent representation, ensuring that the encoder effectively captures the necessary features from the input data. The dimensions $S$ and $E$ are user-defined hyperparameters, which should be empirically determined to optimize the model’s performance. 

\subsubsection{Latent Space}

The latent space component plays a crucial role in our model architecture. In this stage, we utilize the vector quantization mechanism from the VQ-VAE model to quantize the latent space, transforming the continuous latent variables into a set of discrete vectors. This process allows us to discretize the latent space effectively, facilitating the mapping of each $x_s^{t_s}$ to a specific sequence in the quantized space.

\subsubsection{Decoder}

The decoder is tasked with decoding the latent data. Throughout our experimentation, we observed a significant challenge: the original Transformer decoder tended to neglect the latent representations, $z_{q}(x_s^{t_s})$, predominantly relying on the input sequence provided to it. This behavior led to a reduced efficacy in incorporating the latent information into the decoding process. To address this issue, we utilized a module identical to the Transformer's encoder and FNNs to transform the data back to its original shape. This ensures effective utilization of the latent information by the decoder. Finally, the loss function implemented in our model is identical to the one used in the VQ-VAE \cite{VQVAE} model.

\subsection{Prior Inference}

After the training phase, our model is adept at generating a sequence of states for a specific time window. The subsequent step involves modeling the sequence $z_{q}(x_{s}^{t_s})$ as a transition matrix $P_{s}^{t_s} \in \mathbb{R}^{N_e \times N_e}$, where $s$ serves as the system identifier and $t_s$ as the time window identifier. By analyzing the sequence $z_{q}(x_{s}^{t_s})$, it becomes feasible to estimate the corresponding transition matrix $P_{s}^{t_s}$. Each cell in the matrix $P_{s}^{t_s}$ is denoted as ${p_{s}^{t_s}}_{a,b}$ and corresponds to a transition probability from state $a$ to state $b$. It can be calculated as follows:
\begin{equation*}
{p_{s}^{t_s}}_{a,b} = \frac{1}{n} \sum_{m=1}^{S} I({X_{s}^{t_s}}_{m} = a \land {X_{s}^{t_s}}_{m+1} = b),
\end{equation*}
where $I$ is the indicator function that returns one if the transition from $a$ to $b$ occurred, or zero otherwise, $n$ is the number of visits to state $a$, and $S$ represents the number of sequences in the latent space.

To ensure the stability and gradual evolution of the transition matrix $P_{s}^{t_s}$ over the different time windows for a specific system $s$, we employ an exponential moving average (EMA) technique. This smooths out short-term fluctuations while emphasizing longer-term trends in the sequence of states. This approach is particularly effective in capturing dynamic changes within the data across different time windows, leading to a more robust and reliable representation of the transition dynamics in the latent space. The EMA of the transition matrices is given by:
\begin{equation*}
M_{s}^{t_s} = \lambda M_{s}^{t_s-1} + (1-\lambda) P_{s}^{t_s},
\end{equation*}
where $\lambda$ is a smoothing factor. A small value of $\lambda$ will give more weight to most recent observations, making the estimates more responsive to changes while also introducing greater volatility. On the other hand, a high value of $\lambda$ will give more weight to past observations, making the estimates more stable but less responsive to changes.

Finally, to estimate a prior for a system $s$ and time window $t$ we calculate $\pi^{t_s}_s$ that is the steady-state probability of the transition matrix $M_s^{t_s}$. $\pi^{t_s}_s$ is a vector of size $N_{e}$ where each element, ${\pi^{t_s}_s}^a$, is the proportion of times that state $a$ is visited in a random walk in $M_{s}^{t_s}$. This means that ${\pi^{t_s}_s}^a$ reflects the likelihood of the system being in state $a$ after it has reached equilibrium \cite{markov_mixing}. It is important to note that in this work the transition matrices are irreducible, meaning that every state in the state space is accessible from any other state, either directly or indirectly. This is achieved by adding an $\epsilon$ probability to all states that do not have transitions and then normalizing the matrix.

\subsection{Data Similarity}

To infer data similarity, we calculate the $JS$ divergence across different systems under varying time windows. For any given system $j$ at time window $t_j$, the most similar system at a particular time window is identified by finding the one with the minimal $JS$ divergence. This process can be represented as:
\begin{equation*}
    \argmin_{D_s \in D, t_s \in D_s} JS(\pi_j^{t_j}\| \pi_s^{t_s}),
\end{equation*}
where $j \neq s$. During the testing phase, we construct a library of steady-state probability distributions $\pi^{t_s}_s$ generated from the training data. These distributions are then compared with the system currently under study to identify the most similar systems. This process is analogous to finding the k-nearest neighbors, where the distance metric is the $JS$ divergence.

As we possess knowledge of the RUL for the systems in the training set, we can infer the RUL for the system under study. This inference is achieved by selecting the systems, $\pi_s^{t_s}$ that exhibit the lowest $JS$ divergence, denoted as $S_{\pi}$. The total of number of systems selected, $|S_{\pi}|$, is a user-defined parameter that is empirically tuned. The RUL for the system under study is then estimated by averaging the RULs of the systems in $S_{\pi}$. This method leverages the similarity in degradation patterns between the system under study and those in the training set to formulate an informed prediction about the RUL. The process of mapping the steady-state probability distribution $ \pi^{t_j}_j$ to the estimated RUL is represented as follows:

\begin{equation*}
    \hat{RUL^{t_s}_{j}} = \frac{\sum_{s_{\pi} \in S_{\pi}} RUL_{s_{\pi}}}{|S_{\pi}|}
\end{equation*}

\section{Experimentation}

\subsection{CMAPPS Turbofan Dataset}

The experimentation in this work is carried out using the Turbofan Dataset \cite{cmapps} that contains information related to sensors in aircraft. It is important to note that the data in this dataset is not obtained from actual airplane systems. Instead, it is generated using a simulation software called Commercial Modular Aero-Propulsion System Simulation (C-MAPPS) \cite{cmapps}, which accurately simulates the behavior of a turbofan engine. This data closely mirrors real-life aircraft sensor readings, offering a realistic and valuable resource for aircraft systems research and analysis.

The Turbofan dataset is composed of four sub-datasets. Each dataset has a time series readings of 26 features which include: operational settings, system number, time indicator and sensors' values regarding the turbofan engine components. The four different datasets are identical in structure but differ in the number of operational settings and failure modes. There are six different combinations of operational settings which have different impacts on the system's degradation. Each dataset has a training and a testing set. The training sets are run to failure, i.e complete lifetime, whereas the testing sets are pruned some time prior to the failure. In addition, the testing set has information containing the value of the RUL at the time that the system was pruned prior to a failure.


\subsection{Data Preprocessing}

Among the 26 features available in our dataset, 21 are sensor readings. However, not all of these sensor readings are informative for estimating the RUL of the engines. Some sensors exhibit almost constant outputs throughout the engine's lifetime, offering limited predictive value. Consequently, following the guidelines and insights from \cite{HDNN_Al}, we have selectively chosen a subset of 14 sensor readings that are deemed to be more relevant and informative for RUL estimation. In addition to these sensor readings, the operational settings of the engines are also incorporated into the model as input features. Furthermore, we normalize the data using a min-max scaler. 

It is important to note that in this dataset, the RUL is often assumed to follow a linear degradation pattern, starting at a high value and decreasing over time across the time series. However, in reality, a system generally remains healthy during its initial operational phase, with the degradation typically accelerating as it nears the end of its life cycle. Hence, we adopt a piece-wise linear RUL model as suggested in \cite{CNN_Babu, MS_DCNN_Li, LSTM_FNN_Zheng, DCNN_Li, HDNN_Al, MTSTAN_Li, RVE_Costa, LSTM_MLSA_Xia} that helps preventing the overestimation of the RUL, a critical factor in ensuring the reliability of our predictions. In our experimentation, we set the maximum value of the RUL to 125.

\subsection{Experimentation Details}

For our experiments, the encoder of our model consists of three encoding layers, each equipped with three attention heads. The decoder contains two decoding layers, each with three attention heads. It is important to highlight that, based on empirical results, our model demonstrated better performance when trained to predict the RUL directly, as opposed to reconstructing $x$. This improved efficacy in direct RUL prediction is likely due to the relatively weaker prior estimation model in our framework. By focusing on RUL prediction, we effectively reduce the variance in our model's output, leading to more accurate and reliable results. 

Regarding the training conditions for all models, we adhered to the following setup. The ADAM optimizer \cite{Adam} has a learning rate of 0.0002 for all datasets. For the FD001 and FD003 datasets, the models underwent training for $100$ epochs, with each batch comprising $100$ data points and a time window size set to $20$. On the other hand, the FD002 and FD004 datasets necessitated longer training periods, which were set at $125$ and $150$ epochs, respectively. For these datasets, we employed a larger batch size of $256$ and a shorter time window of 10.

The number of sequences in the latent space, $S$, was fixed to 10 for all datasets. The dimension of the latent space, $E$, was set to 32 for all datasets. The number of codebook elements, $N_e$, was set to $25$ for the FD001 and FD003 datasets, $45$ for the FD002 dataset and $60$ the FD004 dataset. The $\lambda$ value of the EMA in the transition matrix estimation was set to $0.9$ for the FD001 and FD003 datasets, and $0.99$ for the FD002 and FD004 datasets. To conclude, the total of number of systems to estimate the RUL, $|S_{\pi}|$, was set to $30$.

\subsection{Evaluation Metrics}


In order to evaluate and compare the performance of our proposed method with other existing techniques, we employed two metrics: the Root Mean Square Error (\textit{RMSE}) and the \textit{Score} function.

\textit{RMSE} is a widely recognized metric for quantifying the deviation between predicted and actual values, particularly suitable in the context of RUL prediction. The \textit{RMSE} is formulated as:
$$RMSE = \sqrt{\frac{1}{N_{te}} \sum_{i=1}^{N_{te}} h_i^2},$$
where $h_i = \hat{RUL_i} - RUL_i$, $\hat{RUL_i}$ is the predicted RUL, and $RUL_i$ is the actual RUL for each test sample and $N_{te}$ is the number of testing samples. The \textit{Score} function was proposed by the International Conference on
Prognostics and Health Management (PHM08) Data Challenge and is given by:

$$Score = \begin{cases}
    \sum_{i=1}^{N_{te}} \euler^{-\frac{h_i}{13}} - 1 \text{ for $h_i < 0$} \\
    \sum_{i=1}^{N_{te}} \euler^{\frac{h_i}{10}} - 1 \text{ for $h_i \geq 0$} \\
\end{cases}, $$

The \textit{Score} function is asymmetrically designed to impose a heavier penalty on late predictions (where the estimated RUL exceeds the actual RUL) compared to early predictions (where the estimated RUL falls short 
of the actual RUL). 

\subsection{Results}

In this section, we present a comparison of our model's performance to several state-of-the-art methods, focusing on the testing dataset. The results of this comparison are detailed in Table \ref{table:results}, which illustrates the performance metrics across different methods.

For the FD001 and FD003 datasets, our model's performance is competitive when compared to other state-of-the-art methods. However, it is important to note that our model did not surpass these methods in terms of performance. This outcome is likely attributable to the relatively smaller size of the FD001 and FD003 datasets. The limited number of samples in these datasets poses a greater challenge in finding similar systems for accurate RUL prediction. In contrast, for the FD002 and FD004 datasets, our model demonstrates a significant improvement in performance. With respect to the \textit{RMSE} we observe an increase in performance of approximately 7\% and 19\% over the best performing model (the underlined value in Table \ref{table:results}) for the FD002 and FD004 datasets, respectively. This enhanced performance on the larger and more complex FD002 and FD004 datasets indicates the efficacy of our model in handling datasets with a broader range of samples and more varied conditions.

\begin{table}[ht]
\centering
\caption{Performance comparison of state-of-the art methods with the proposed method for the testing set systems (the best result for each dataset is emphasized in bold).}
\setlength{\tabcolsep}{4pt}
\begin{tabular}{cccccc}
    \toprule
    \multirow{2}{*}{Algorithm} & \multirow{2}{*}{Metric} & \multicolumn{4}{c}{Dataset} \\
    \cline{3-6}
    & & FD001 & FD002 & FD003 & FD004 \\ 
    \midrule
    \multirow{2}{*}{\centering \texttt{Proposed Method}}
    & RMSE & 12.61 & \textbf{13.05} & 12.71 & \textbf{13.99} \\
    & Score & 245.65 & 917.09 & 364.57 & \textbf{892.79} \\
    \midrule
    \multirow{2}{*}{\centering \texttt{CNN \cite{CNN_Babu}}}
    & RMSE & 15.84 & 30.29 & 13.53 & 29.17 \\
    & Score & 374.65 & 13570 & 318.93 & 7886.40 \\
    \midrule
    \multirow{2}{*}{\centering \texttt{DCNN \cite{DCNN_Li}}}
    & RMSE & 12.61 & 22.36 & 12.64 & 23.31 \\
    & Score & 273.7 & 10412 & 284.1 & 12466 \\
    \midrule
    \multirow{2}{*}{\centering \texttt{LSTM-FNN \cite{LSTM_FNN_Zheng}}}
    & RMSE & 16.14 & 24.49 & 16.18 & 28.17 \\
    & Score & 338 & 4450 & 852 & 5550 \\
    \midrule
    \multirow{2}{*}{\centering \texttt{MS-DCNN \cite{MS_DCNN_Li}}}
    & RMSE & 11.44 & 19.35 & 11.67 & 22.22 \\
    & Score & 196. 22 & 3747 & 241.89 & 4844 \\
    \midrule
    \multirow{2}{*}{\centering \texttt{HDNN \cite{HDNN_Al}}}
    & RMSE & 13.02 & 15.24 & 12.22 & 18.16 \\
    & Score & 245 & 1282.42 & 287.72 & 1527.42 \\
    \midrule
    \multirow{2}{*}{\centering \texttt{MTSTAN \cite{MTSTAN_Li}}}
    & RMSE & \textbf{10.97} & 16.81 & \textbf{10.90} & 18.85 \\
    & Score & \textbf{175.36} & 1154.36 & \textbf{188.22} & 1446.29 \\
    \midrule
    \multirow{2}{*}{\centering \texttt{RVE \cite{RVE_Costa}}}
    & RMSE & 13.42 & 14.92 & 12.51 & \underline{16.37} \\
    & Score & 323.82 & 1379.17 & 256.36 & 1845.99 \\
    \midrule
    \multirow{2}{*}{\centering \texttt{LSTM-MLSA \cite{LSTM_MLSA_Xia}}}
    & RMSE & 11.57 & \underline{14.02} & 12.13 & 17.21 \\
    & Score & 252.86 & \textbf{899.18} & 370.39 & 1558.48 \\
    \bottomrule
\end{tabular}
\label{table:results}
\end{table}

\begin{figure}[h!] 
  \begin{subfigure}[b]{0.5\linewidth}
    \centering
    \includegraphics[width=\linewidth]{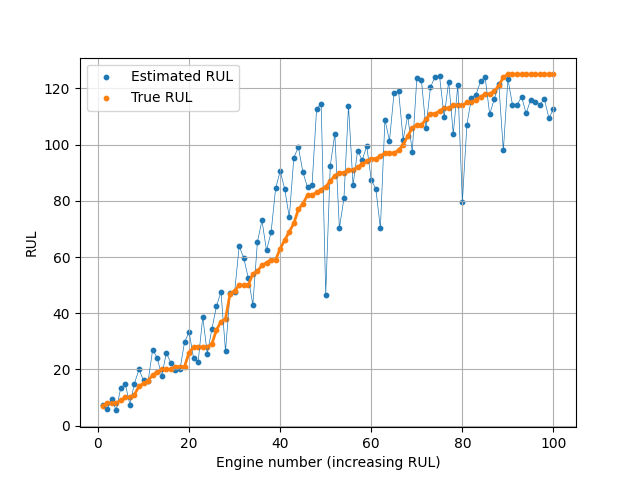} 
    \caption{FD001} 
    \label{fig7:a} 
    \vspace{4ex}
  \end{subfigure}
  \begin{subfigure}[b]{0.5\linewidth}
    \centering
    \includegraphics[width=\linewidth]{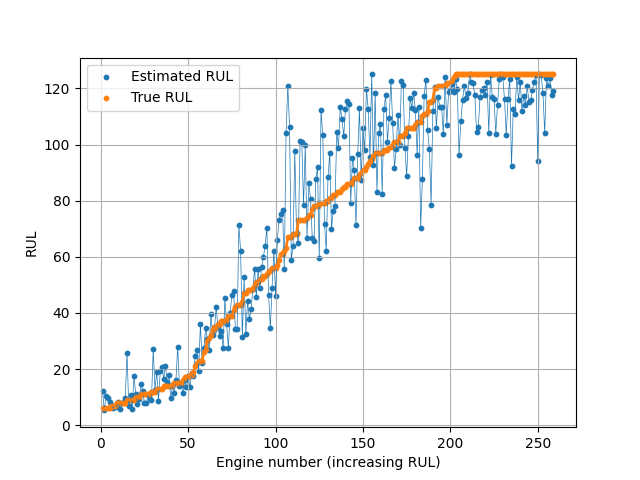} 
    \caption{FD002} 
    \label{fig7:b} 
    \vspace{4ex}
  \end{subfigure} 
  \begin{subfigure}[b]{0.5\linewidth}
    \centering
    \includegraphics[width=\linewidth]{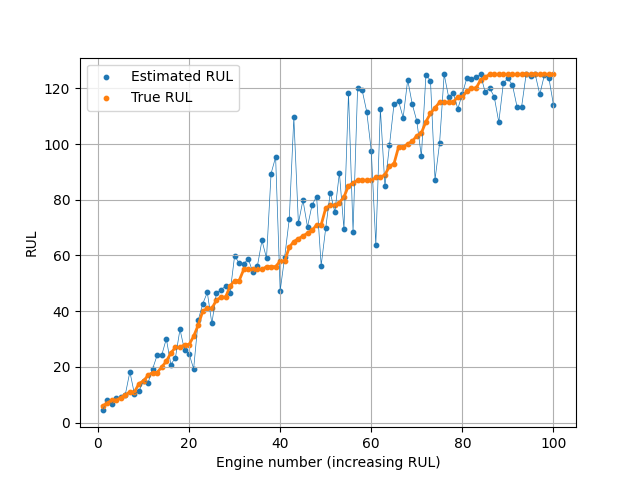} 
    \caption{FD003} 
    \label{fig7:c} 
  \end{subfigure}
  \begin{subfigure}[b]{0.5\linewidth}
    \centering
    \includegraphics[width=\linewidth]{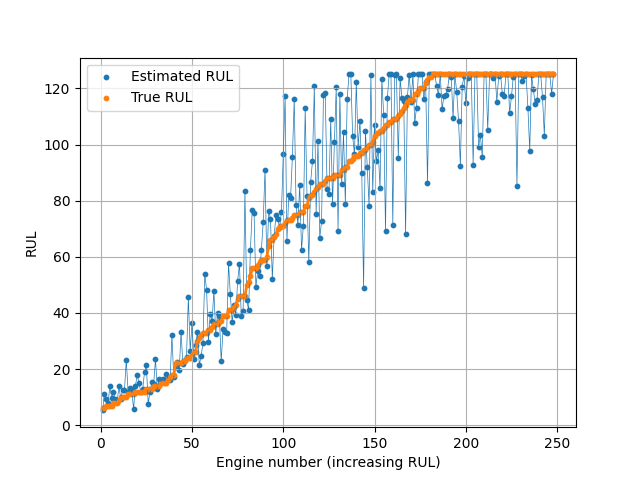} 
    \caption{FD004} 
    \label{fig7:d} 
  \end{subfigure} 
  \caption{The results of the RUL predictions for the C-MAPSS testing dataset. (a) Prediction for the 100 testing systems in FD001. (b) Prediction for the 256 testing systems in FD002. (c) Prediction for the 100 testing systems in FD003. (d) Prediction for the 248 testing systems in FD004. In each subplot, the actual RULs are depicted by orange lines, while the estimated RULs are shown in blue.}
  \label{fig:graphs} 
\end{figure}

To summarize our findings, Figure \ref{fig:graphs} presents a comparison between the model's RUL predictions and the actual RUL across all datasets. For enhanced visualization, the testing systems are arranged in ascending order of their actual RUL, meaning that the RUL values increase from left to right on the graph. It can be observed that the predicted RUL values closely align with the ground truth, especially for engines with smaller RUL. This observation is particularly significant, as engines with a smaller RUL are closer to potential failure. Achieving higher accuracy in these cases is crucial for the effective implementation of CBM actions at optimal times, thereby avoiding catastrophic failures. The higher accuracy for engines with smaller RUL can be attributed to the fact that during the early stages of prediction the system prior estimation is relatively weak, meaning that the distinction between different systems' data is not as pronounced. As the systems approach their end-of-life and the degradation becomes more evident, the model is able to make more accurate predictions. This underscores the effectiveness of our approach in identifying and responding to critical degradation patterns, particularly in the latter stages of a system's lifecycle.

\section{Conclusion}

In this paper, we proposed a novel approach for RUL estimation, a crucial aspect of CBM. Our methodology integrates a Transformer-based architecture with vector quantization model for RUL prediction. This framework extends beyond mere direct estimation of the RUL, emphasizing the significance of latent data divergence as a key factor in predicting system degradation. Our approach represents a shift from traditional predictive models, directing attention towards a more nuanced understanding of the latent features and their variations over time.

The results of our implementation demonstrate substantial efficacy, particularly within larger datasets (FD002 and FD004), where the model exhibited performance improvements of around 7\% and 19\% respectively over baseline models. These results highlight the capability of our approach in capturing and interpreting complex data that is characteristic of CBM scenarios. However, the performance on smaller datasets (FD001 and FD003), though competitive, did not surpass existing state-of-the-art methods. This outcome suggests potential areas for refinement in our approach, particularly in its application to datasets with fewer samples by creating a better prior estimate.

Our research opens up several promising directions for future work. Primarily, there is an opportunity to refine and enhance the architecture to improve its performance. This could involve more sophisticated handling of latent data divergence or integrating additional components to enrich the model's understanding. Another interesting avenue is extending the application of our methodology to other domains, exploring its adaptability and effectiveness in various data contexts. 

\bibliographystyle{IEEEtran}
\bibliography{literature}

\end{document}